\documentclass[sigconf]{acmart}

\usepackage{graphicx}
\usepackage{tikz}
\usetikzlibrary{tikzmark, arrows.meta}
\usepackage{xcolor}
\usepackage[list=true]{subcaption}
\usepackage{textcomp}
\usepackage{xspace}
\usepackage{todonotes}

\usepackage{physics}
\usepackage{sistyle}
\SIthousandsep{,}
\usepackage{array, tabularx, tabulary, multirow}
\newcolumntype{Y}{>{\centering\arraybackslash}X}
\newcolumntype{P}[1]{>{\centering\arraybackslash}p{#1}}
\newcolumntype{M}[1]{>{\centering\arraybackslash}m{#1}}

\usepackage{array}
\usepackage{threeparttable}
\usepackage{tabularx}
\usepackage{makecell}
\usepackage{threeparttablex}
\usepackage{pifont}
\usepackage[tableposition=top]{caption}
\definecolor{lightergray}{gray}{0.95}
\begin{document}

\title{A Real-time Human Pose Estimation Approach for Optimal Sensor Placement in Sensor-based Human Activity Recognition}

\author{Orhan Konak}
\email{orhan.konak@hpi.de}
\orcid{0000-0003-1884-8029}
\affiliation{
  \institution{Hasso Plattner Institute\\University of Potsdam}
  \streetaddress{Prof.-Dr.-Helmert-Straße 2-3}
  \city{Potsdam}
  \country{Germany}
  \postcode{14482}
}

\author{Alexander Wischmann}
\email{alexander.wischmann@student.hpi.uni-potsdam.de}
\affiliation{
  \institution{Hasso Plattner Institute\\University of Potsdam}
  \streetaddress{Prof.-Dr.-Helmert-Straße 2-3}
  \city{Potsdam}
  \country{Germany}
  \postcode{14482}
}

\author{Robin van de Water}
\email{robin.vandewater@hpi.de}
\affiliation{
  \institution{Hasso Plattner Institute\\University of Potsdam}
  \streetaddress{Prof.-Dr.-Helmert-Straße 2-3}
  \city{Potsdam}
  \country{Germany}
  \postcode{14482}
}

\author{Bert Arnrich}
\email{bert.arnrich@hpi.de}
\affiliation{%
  \institution{Hasso Plattner Institute\\University of Potsdam}
  \streetaddress{Prof.-Dr.-Helmert-Straße 2-3}
  \city{Potsdam}
  \country{Germany}
  \postcode{14482}
}

\begin{abstract}
Sensor-based Human Activity Recognition facilitates unobtrusive monitoring of human movements. However, determining the most effective sensor placement for optimal classification performance remains challenging. This paper introduces a novel methodology to resolve this issue, using real-time 2D pose estimations derived from video recordings of target activities. The derived skeleton data provides a unique strategy for identifying the optimal sensor location. We validate our approach through a feasibility study, applying inertial sensors to monitor 13 different activities across ten subjects. Our findings indicate that the vision-based method for sensor placement offers comparable results to the conventional deep learning approach, demonstrating its efficacy.
This research significantly advances the field of Human Activity Recognition by providing a lightweight, on-device solution for determining the optimal sensor placement, thereby enhancing data anonymization and supporting a multimodal classification approach. 
\end{abstract}





\maketitle

\begin{figure}[htbp]
\centerline{\includegraphics[width=3.7cm]{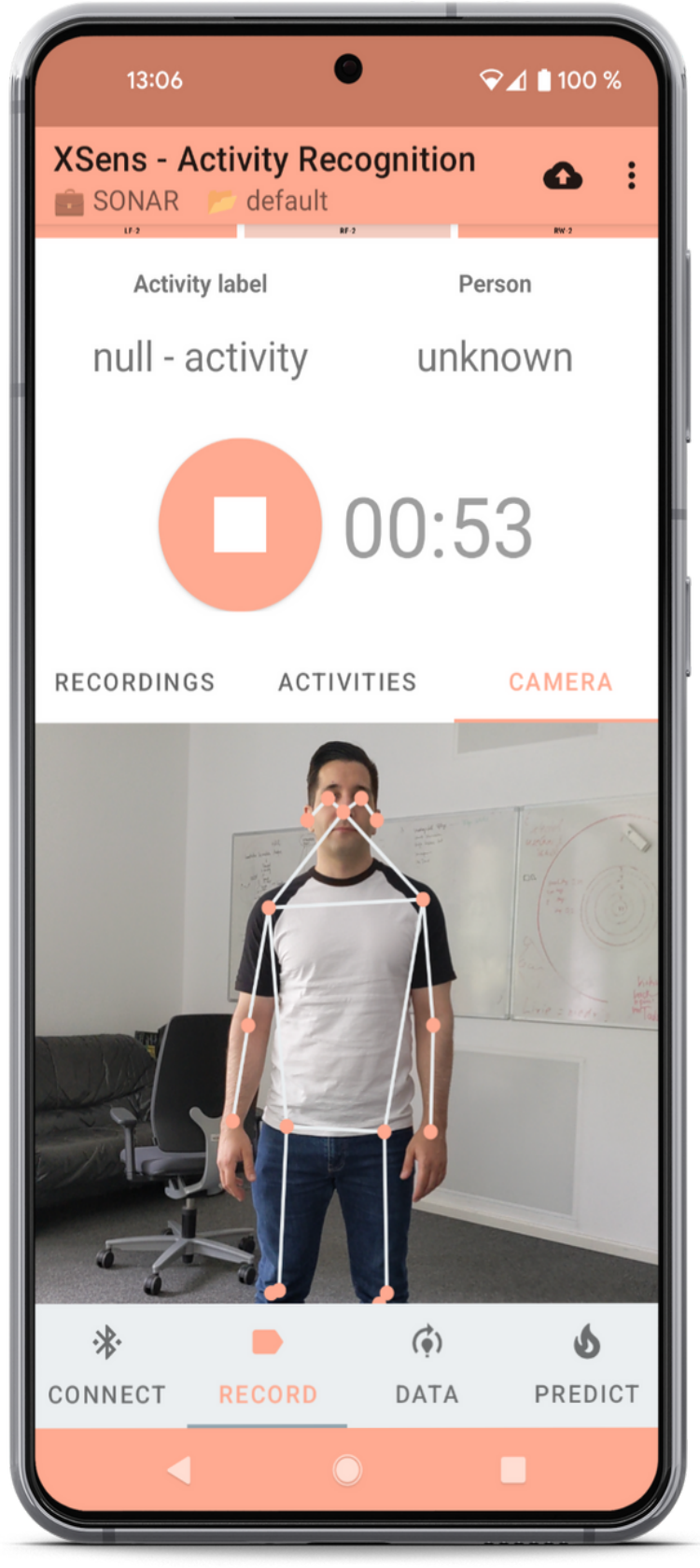}}
\caption{Optimal sensor placement through real-time pose estimation.}
\label{fig:optimal_sensor_placement}
\end{figure}

\section{Introduction}
\label{sec:introduction}

Sensor-based Human Activity Recognition (HAR), as part of pervasive computing, describes the process of distinguishing movements by using Inertial Measurement Units (IMU). IMUs primarily measure quantities such as acceleration and angular velocity. 
Depending on the performed movements, the data constitute distinct time series patterns, which can be classified by a Machine Learning (ML) model.
The movements can range from low-level activities, such as walking and standing, to high-level activities, which are combinations of multiple low-level activities. HAR has potential applications in various domains, including healthcare, sports, and smart environments~\cite{serpush2022wearable, sport_HAR, 10.1145/3550294}. However, challenges remain regarding the placement of sensors to achieve higher classification performance and the preservation of privacy~\cite{SenseCollect}.

Optimal placement of on-body sensors is a major challenge in HAR, as the location of these sensors directly influences the activity classification~\cite{kunze2014sensor}. Therefore, many sensors or experiences from similar studies are consulted before conducting an own study. Data acquisition and labeling in HAR are another challenge and are typically facilitated by the use of a camera, introducing privacy concerns. Video recordings in privacy-sensitive areas often raise ethical questions, thus prompting convoluted workarounds~\cite{SenseCollect}. To address these challenges, we introduce a method designed to optimize the sensor placement while preserving privacy in HAR.

To preserve privacy, we convert video data into real-time 2D human pose estimations, creating a skeleton representation of the subject's movements, as depicted in \autoref{fig:optimal_sensor_placement}. These 2D keypoints not only aid in recommending optimal sensor placement for given activities but also enrich the classification process in a multimodal classification approach. 

To evaluate the effectiveness of our approach, we conducted a lab study with ten subjects performing nursing activities.
These activities were recorded over eight hours, providing an open access rich dataset that encompassed a wide range of movements and scenarios. We evaluated our method with nursing activities due to their complexity, i.e., the variety of low-level and high-level tasks involved and the relevance to healthcare, one of the main domains where HAR can have a significant impact~\cite{10.1145/3351244}. The evaluation allowed us to assess our approach's performance in terms of sensor placement, multimodal classification, as well as its ability to preserve privacy during these processes. Through our evaluation, we found that multimodality increased F1 score by up to 4.4\%. Furthermore, three out of four sensor placement suggestions were equal to the best-performing deep learning model, a CNN-LSTM, with an overall Kendall’s tau of 0.8. Therefore, our research contributes to the field of HAR through an on-device method using 2D pose estimation for determining optimal sensor placement, requiring only 500 data points. This approach can even work with publicly available video footage of target activities. Furthermore, the utilization of 2D keypoints from pose estimation not only enhances privacy during data collection but also facilitates a multimodal approach to HAR, creating an efficient fusion between IMUs and 2D keypoints.

The remainder of the paper is structured as follows: In Section~\ref{sec:related_work}, we contextualize our research within existing approaches of optimal sensor placement and multimodality. In Section~\ref{sec:app_architecture}, we provide details on our approach. In Section~\ref{sec:pilot_study}, we uncover and assess the practicality of the proposed features through a feasibility study on nursing activities. Section~\ref{sec:discussion} discusses the results and limitations of our study, while Section~\ref{sec:conclusion} concludes the paper and outlines potential future research directions.

\section{Related Work}
\label{sec:related_work}

HAR is a field that has seen significant progress in recent years, particularly in the context of sensor-based HAR with wearable sensors~\cite{sensor_based1, sensor_based2, sensor_based3}. HAR has many potential applications, e.g., healthcare, fitness, security, and surveillance. In healthcare, HAR can be used to monitor the activity levels of elderly patients with chronic diseases~\cite{har_healthcare1, har_healthcare2}. In fitness, HAR can be used to track physical activity and provide feedback to athletes~\cite{har_fitness1, har_fitness2}. In security and surveillance, HAR can be used to monitor the activities of people in restricted areas or identify potential threats~\cite{har_security1}. As such, a vast body of literature encompasses a range of research sub-areas, including sensor placement optimization and a multimodal classification approach.
Throughout this section, we highlight the strengths and limitations of the existing approaches and compare them to our approach.

\subsection{Sensor Placement in HAR}

Conducting a study on sensor-based HAR with IMUs necessitates the question of sensor placement. The classification result highly depends on the incoming data, which varies with the location and number of used sensors for different body parts~\cite{kunze2014sensor}. Research suggests that the most accurate results are achieved when sensors are positioned at the chest, ankles, and thighs~\cite{serpush2022wearable}. Evidence indicates that harnessing accelerometers on both the upper and lower torso concurrently can significantly enhance the precision of activity recognition~\cite{keyvanpour2019eslmt, fu2020sensing}. \citet{sensor_placement_number} compared the performance of different placements of accelerometer devices on the body in categorizing physical activities and estimating energy expenditure in older adults. They used five different body positions for accelerometer placement: wrist, hip, ankle, upper arm, and thigh. The study concludes that considering the placement of the accelerometer devices is important in optimizing the accuracy of HAR.
\citet{virtual_sensor_placement} discusses how the performance of HAR systems is affected by the sensor position and proposes an optimization scheme to generate the optimal sensor position from all possible locations given a fixed number of sensors. The system uses virtual sensor data to access the training dataset at a low cost and can help make decisions about sensor position selection with great accuracy using feedback.

In contrast to existing approaches, our approach does not require any sensor setup. Instead, we rely on human pose estimations using either self-recorded videos or existing videos of the target activities to determine the optimal sensor placement. This significantly reduces the setup and calibration efforts required for HAR and eliminates the need for physical sensors. Additionally, our approach involves much less computation compared to classical approaches that involve training and testing with large datasets, making it a more efficient and practical solution for real-world applications.

\subsection{Multimodal Approaches in HAR}
Multimodal HAR has gained more attention in recent years due to its potential to leverage multiple sources of sensory data and provide more accurate and robust activity recognition compared to unimodal approaches~\cite{yadav2021review, 10.1016/j.inffus.2021.11.006}. \citet{10.1145/3377882}, for example, explored methods of fusing and combining multi-representations of sensor data, using data-level, feature-level, and decision-level fusions with Deep Convolutional Neural Networks and achieved promising results. \citet{das2020mmhar} proposed MMHAR-EnsemNet, which uses four different modalities to perform sensor-based HAR and has been evaluated on two standard benchmark datasets.

In contrast to these multimodal approaches, we utilize a single device for collecting data from IMUs and videos. This data is transformed in real-time into 2D human pose estimations, providing an inherently given multimodal datastream for recording and classification.

\section{Methods}
\label{sec:app_architecture}

This section outlines our approach toward the collection and recording of data as well as the proprietary method for determining the optimal sensor placement.

\subsection{Connection}
\label{sub_sec:connection}

We decided to use the Xsens\texttrademark \xspace DOT sensor\footnote{For detailed information see the user manual: \url{https://www.xsens.com/hubfs/Downloads/Manuals/Xsens\%20DOT\%20User\%20Manual.pdf}} as a standalone device at specific on-body locations, which allows for unobtrusive data recording because of its size and weight. We used the Xsens DOT Android software development kit (SDK version v2020.4)\footnote{\href{https://base.xsens.com/s/article/Xsens-DOT-Software-Package?language=en_US}{https://base.xsens.com/s/article/Xsens-DOT-Software-Package?language=en\_US}} to build an app for scanning, connecting, and receiving data in real-time. Xsens DOT uses Bluetooth for data transmission to the host device. Although there is no connection limit in the Xsens DOT SDK services, the central devices' hardware and operating system constraints limit the maximum number of sensors that can be connected simultaneously. Using Android, it is possible to connect up to seven sensors. The output rate for the measurement can be specified and ranges from 1 Hz to 60 Hz for real-time streaming. The recording mode allows up to 120 Hz. All sensors are time-synced after synchronization.
Transmitted data includes calibrated orientation data (quaternion), calibrated inertial data, and magnetic field data.

\subsection{Recording}
\label{sub_sec:recording}

Connecting the IMUs to our application facilitates capturing various sensor data types, including quaternions, free acceleration, angular velocity, and the magnetic field normalized to Earth's field strength, at adjustable output rates. The application also supports video recording. While the output rate can be set according to the user's preference, it is ultimately limited by the device's hardware capabilities. The recorded video is leveraged to generate real-time pose estimations. These estimations serve three primary purposes: they guide the determination of optimal sensor placement, ensure the anonymization of the incoming data stream, and support a multimodal classification approach, thereby enhancing the accuracy and utility of our method.

\subsection*{Optimal Sensor Placement}
\label{subsub_sec:osp}

The optimal sensor placement is derived through 2D pose estimations. Pose estimation is a computer vision technique that refers to detecting humans and their poses from image and video data~\cite{openpose}. We use the incoming video data for real-time pose estimations to create key body joints. To make it work on the device, we use MoveNet Thunder's\footnote{\href{https://tfhub.dev/google/movenet/singlepose/thunder/4}{https://tfhub.dev/google/movenet/singlepose/thunder/4}} pre-trained TensorFlow Lite (TFLite) pose estimation model~\cite{tensorflow2015-whitepaper}. The outcome is a landmark of 17 keypoints in 2D at different body locations, such as ankles, knees, hips, wrists, elbows, shoulders, and some facial parts in each timestamp. Since the position data has a causal link to the acceleration through the second derivative, each keypoint can be understood as an accelerometer. Hence, we interpret each keypoint as a potential location for sensor placement. We implemented an algorithmic procedure to calculate the optimal sensor placement, which works in three phases.

The selected pose estimations underwent preprocessing, involving the combination of keypoints. Not all 17 detected keypoints were suitable for sensor placement, leading to the consolidation of several keypoints. The head-related and hip keypoints were replaced with a single, average keypoint as they are part of one bone segment. To mitigate rapid changes in keypoint coordinates due to movement or incorrect pose estimation, the remaining 12 keypoints were centralized, with their center of mass located at point $(0.5, 0.5)$ in each data series. For comparison with a real-life setting, we reduced the number of keypoints to five by selecting the two wrists, two ankles, and pelvis. \citet{dip-imu} showed in their work on Deep Inertial Poser that these locations contain rich information for full body pose estimation, making them ideal for evaluation purposes. The head was excluded from sensor placement as it was considered less relevant to the movements under study.

In the second step, we define and calculate a cross-validated feature metric $D_k$, inspired by the cosine distance formula. Our formula determines the optimal sensor placement. For each activity, we require a minimum sequence of \num{500} data points in $x$ and $y$, corresponding to a 50 s recording with 10 Hz. The number \num{500} was determined through experimentation with different sequence lengths. Activities with a recording time longer than \num{500} data points are cut to a uniform length of \num{500}. We convert these sequences into a multivariate per-keypoint time series. We denote activities by $a_i \in A={a_1, a_2, \dots, a_n}$, where $n$ represents the number of activities. A concatenated time series is created for each $a_i \in A$, and each of the 12 combinations of $s$ subsets of the keypoints; this results in a vector $A_k^i$ of length $s \times 500$ of two-dimensional data points for each. We hypothesize that more distinct vectors, i.e. a lower dot product, between the activities correspond to more distinct features, thus leading to higher classification accuracy. Therefore, a higher $D_k$ value coincides with a higher likelihood of an optimal sensor location. Using the following expression, we calculate the $D_k$ value for each combination $k$, indicating the difference of the respective keypoint vectors between the different activities:

\begin{align}
    \label{for:cosine_distance}
    D_k:= \sum_{i=1}^{n-1}\sum_{j=i+1}^{n}|1 - \frac{\mathbf{A^i_{k}} \cdot \mathbf{A^j_{k}}}{\|\mathbf{A^i_{k}}\| \|\mathbf{A^j_{k}}\|}|.
\end{align}

Finally, all keypoint combinations are sorted by $D_k$  and displayed in a dialog box. Providing a vision-based virtual sensor approach allows us to find the optimal sensor placement with less effort than using physical IMUs with subsequent model training and evaluation. Therefore, having the sensors at hand and collecting IMU data is not required. An existing or self-recorded video on targeted activities suffices to receive recommendations for the sensor placement.

\section{Experimental Evaluation: Nursing Activity Recognition}
\label{sec:pilot_study}

In order to evaluate the effectiveness of the algorithmic approach, we collected data on nursing activities under the instruction of a real nurse.
Nursing activities were selected as they encompass a wide range of complex and diverse tasks, requiring accurate and efficient data collection and classification. By applying our approach to this real-world scenario, we can effectively demonstrate its capabilities in addressing the challenges of sensor placement, multimodal classification, and privacy preservation.

\subsection{Data Description}
\label{sub_sec:data}

\begin{figure*}
\centering
\subcaptionbox{Null-activity}{\includegraphics[width=0.235\textwidth]{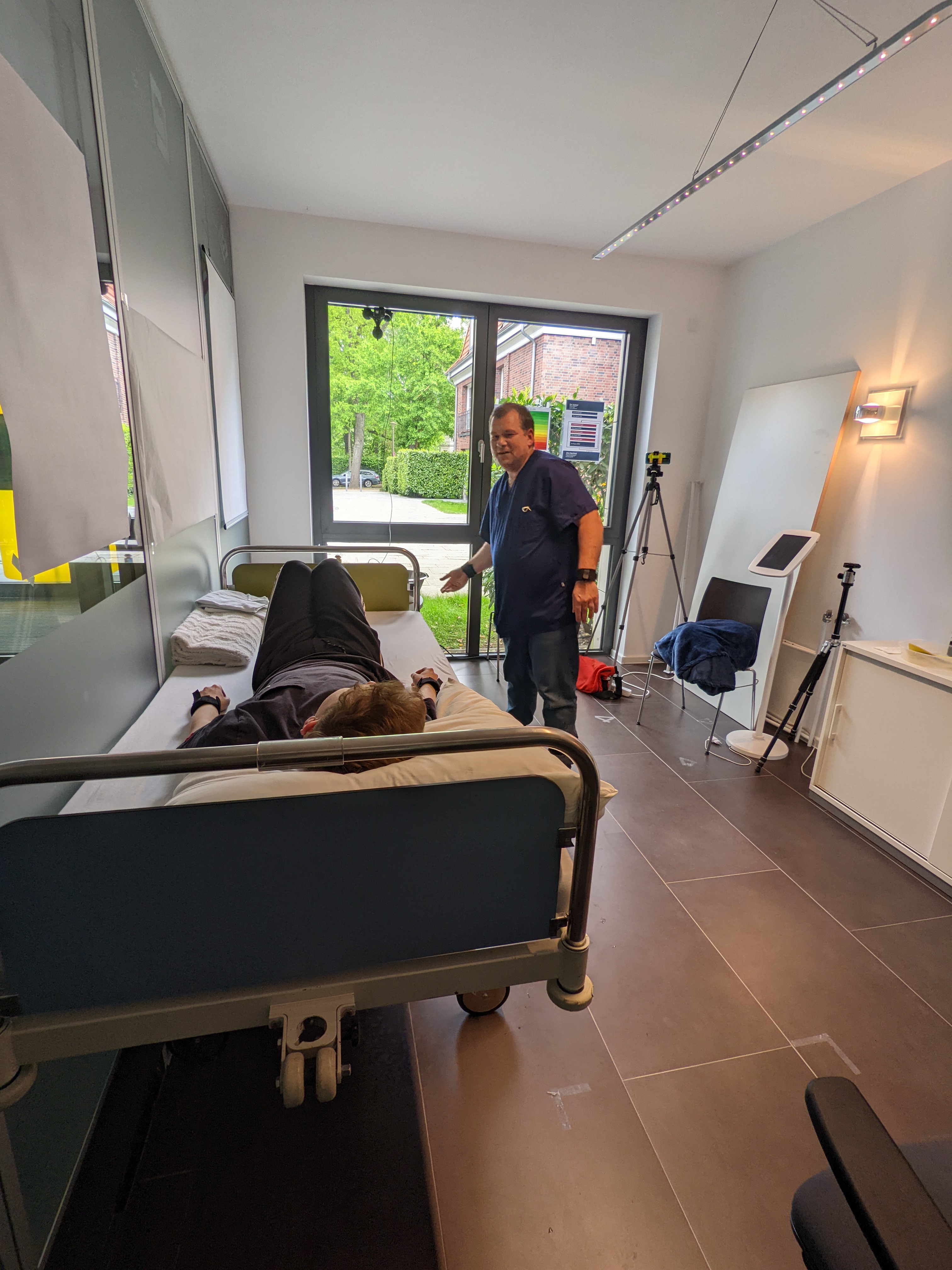}}%
\hfill
\subcaptionbox{Assist in getting dressed}{\includegraphics[width=0.235\textwidth]{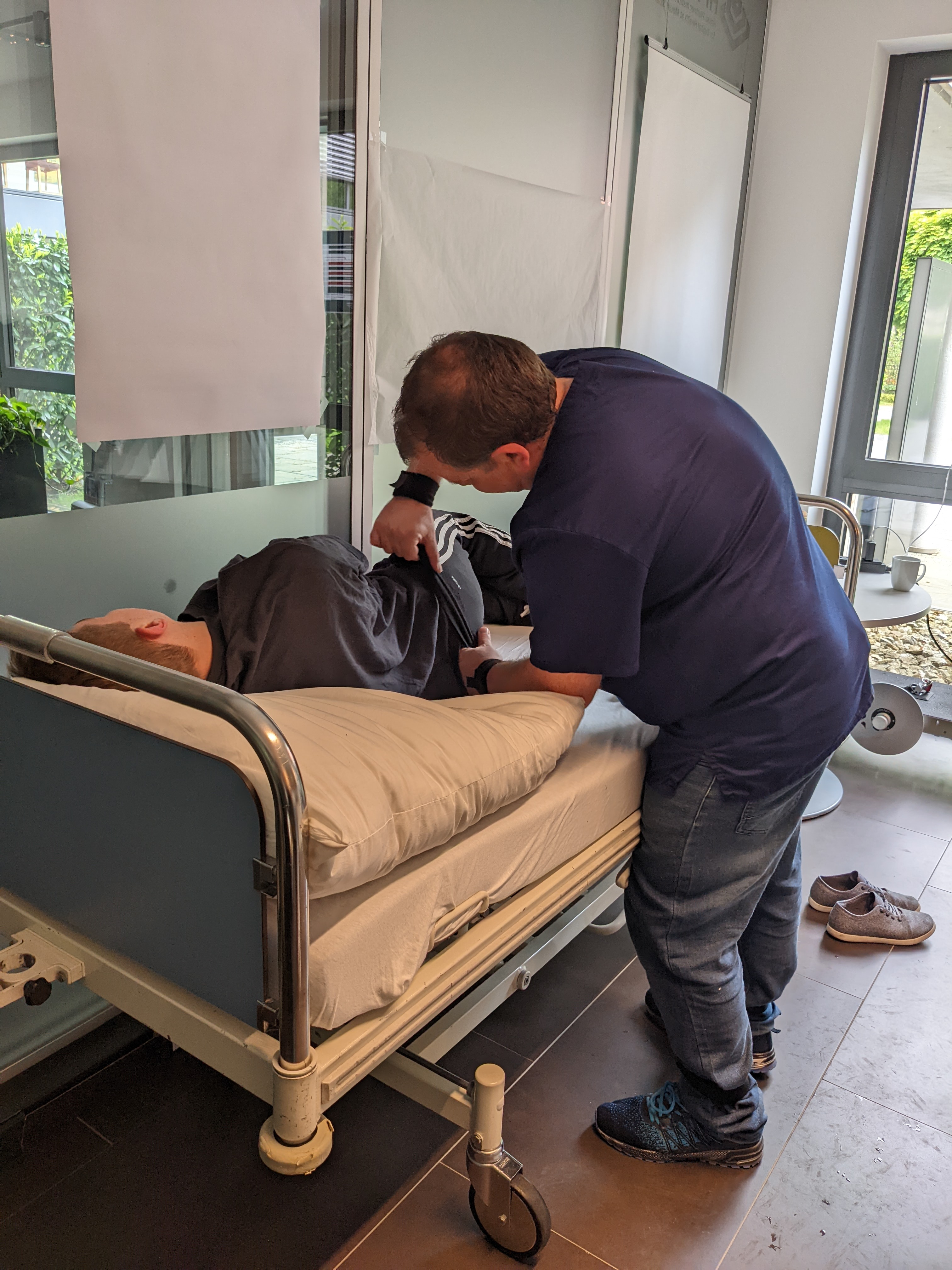}}%
\hfill
\subcaptionbox{Reposition}{\includegraphics[width=0.235\textwidth]{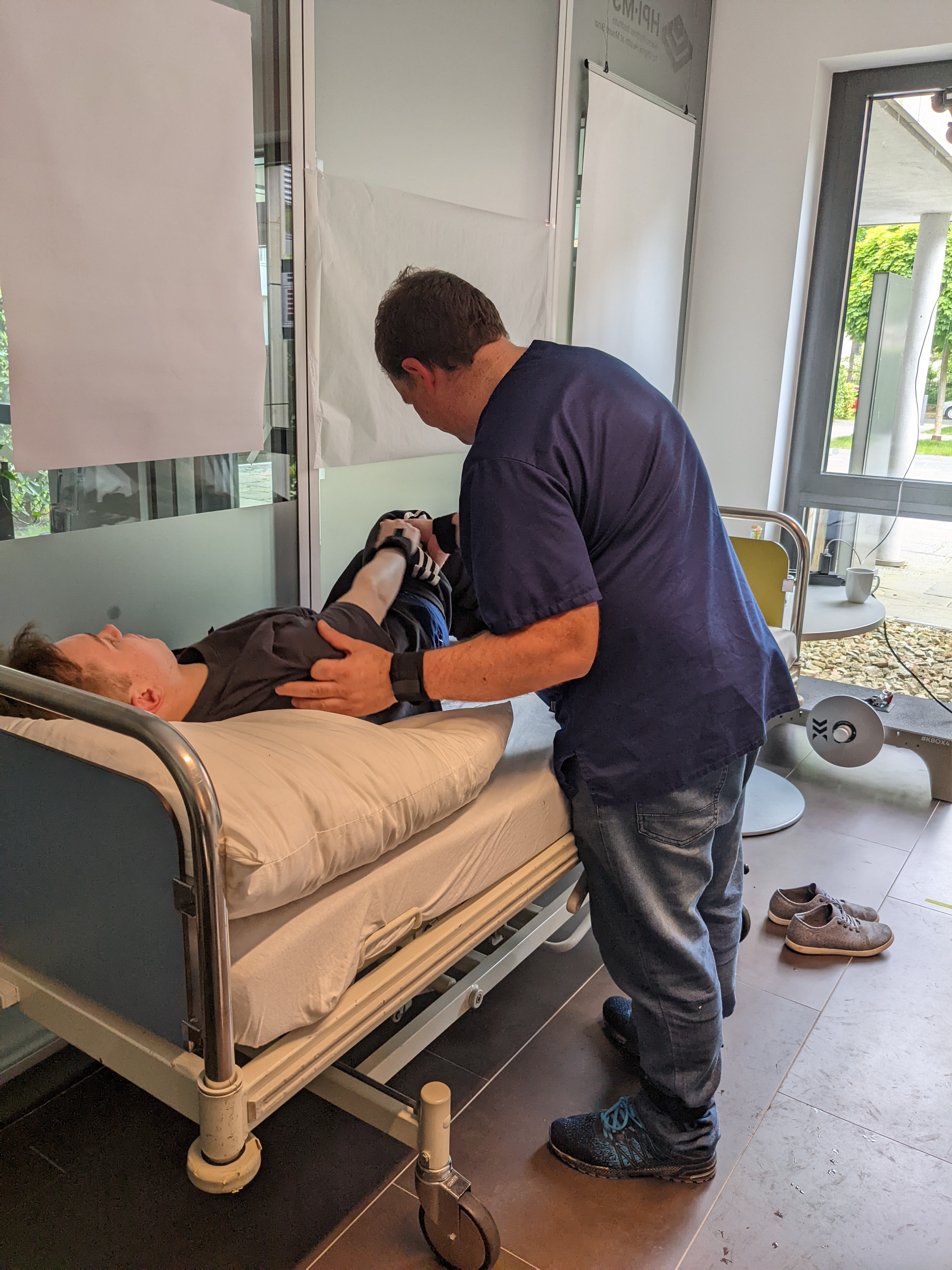}}%
\hfill
\subcaptionbox{Morning care - full body washing in bed}{\includegraphics[width=0.235\textwidth]{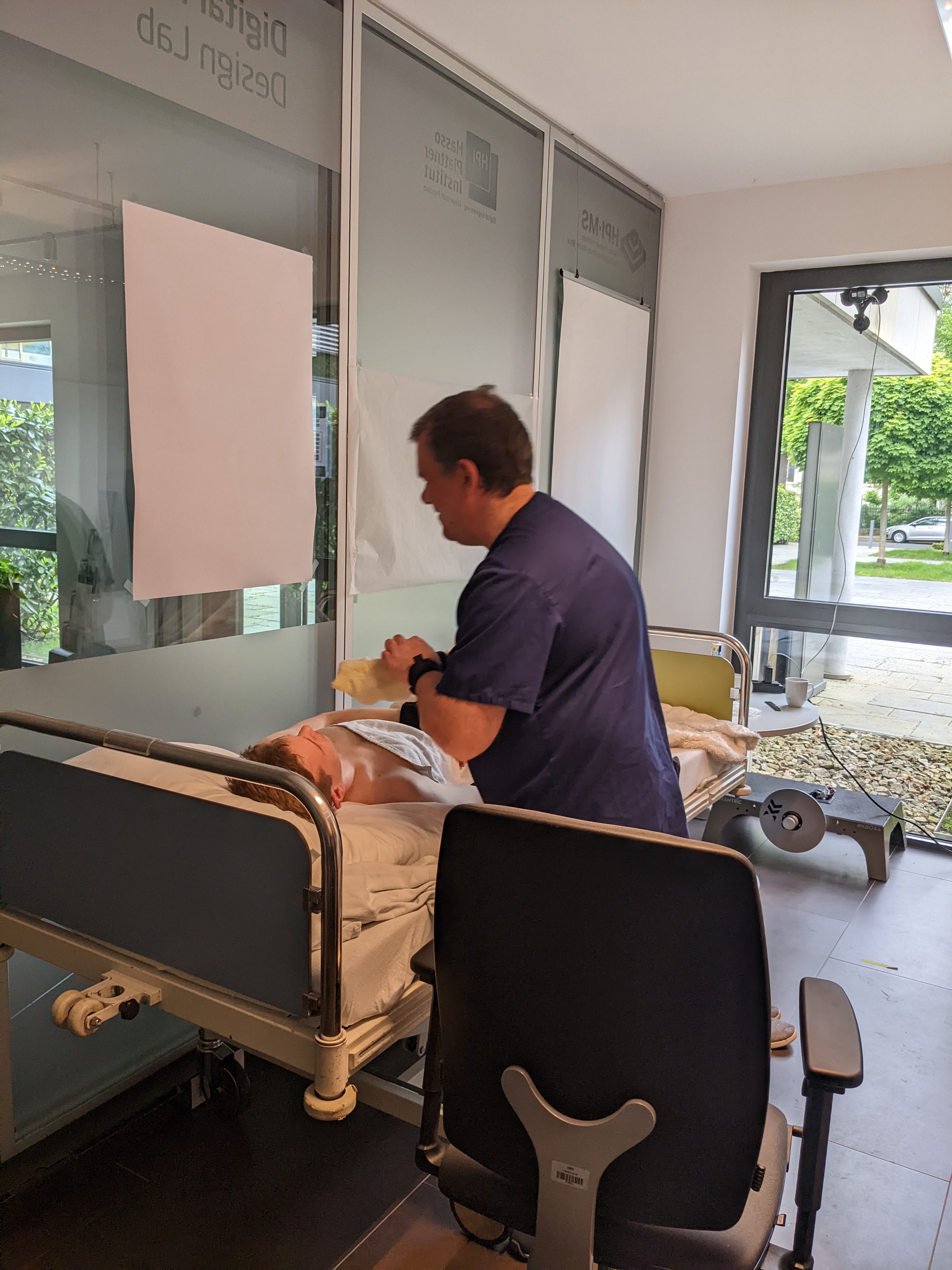}}%
\caption{A demonstration of different nursing activities conducted in a lab study.}
\label{fig:feasibility_study}
\end{figure*}

Data for this study were collected using five Xsens DOT sensors with a 60 Hz output rate at the positions left wrist, right wrist, pelvis, left ankle, and right ankle.
The sensor data outputs, consisting of 14 features, contain four-dimensional quaternion values, four-dimensional angular velocity determined by the derivative of the quaternion values, three-dimensional acceleration values, and three-dimensional magnetic field values.

\begin{figure}[h!]
\centerline{\includegraphics[width=0.83\linewidth]{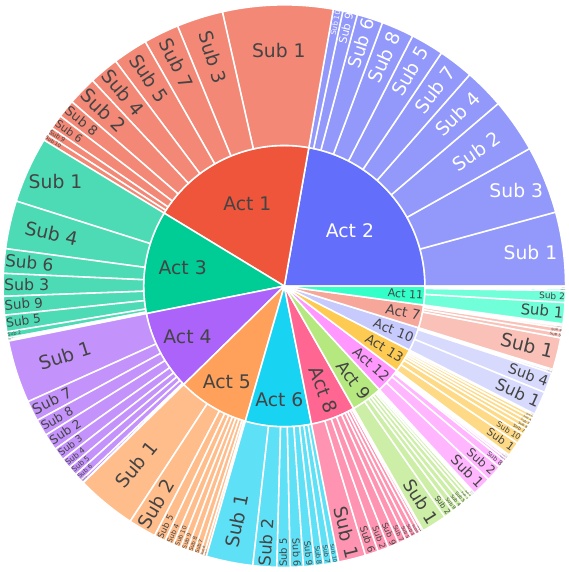}}
\caption{Nested pie chart showing the distribution of the different activities and subjects. \textbf{Legend}: Act 1 = null activity; Act 2 = assist in getting dressed; Act 3 = full body washing in bed; Act 4 = feeding; Act 5 = make the bed; Act 6 = clean up; Act 7 = skin care; Act 8 = push the wheelchair; Act 9 = wheelchair transfer; Act 10 = comb hair; Act 11 = wipe up; Act 12 = prepare medication; Act 13 = serve food; Act 14 = documentation; Sub1-Sub10 = Subjects 1 to 10}
\label{lab_statistics}
\end{figure}

The dataset comprises 13 activities, including ten subjects, leading to 51 recordings with an overall of \num{1519418} data points per feature, which corresponds to 486.8 minutes ($\sim$8 hours) recording. \autoref{fig:feasibility_study} shows an excerpt of the activities conducted in the study. \autoref{lab_statistics} highlights the distribution of the subjects on each activity. We chose to utilize the Xsens DOT sensors for our study due to their accuracy, reliability, and suitability for the healthcare scenario we focused on. These sensors provide high-quality data, which is essential for accurate activity classification involving the 13 specific activities we examined. Although our experiments and comparisons were conducted using Xsens DOT sensors, our findings and insights can be applied to other sensor types and devices. The methodology and techniques we employed for data collection, classification, and privacy preservation are generally applicable to a wide range of HAR scenarios, regardless of the specific sensors used.

\subsection{Results}
\label{sub_sec:results}

We trained a CNN-LSTM deep learning model. We used hyperparameter optimization using grid search~\cite{scikit-learn} for the window length and learning rate. This resulted in a learning rate of \num{1e-4} and an input size of $600 \times 70$. The input size corresponds to a window length of 600 (equal to 10 s with 60 Hz) and 14 features from each sensor ($14\cdot 5=70$). The used model contains a preprocessing step for filling missing values and a batch-normalization layer to standardize the inputs in each feature row. The output of the network is a dense softmax layer with the number of activity classes. We used the Adam\cite{adam} optimization approach. The categorical cross-entropy loss function was used for a multi-class classification problem:

\begin{align*}\label{eq:loss}
    L = -log\left ( \frac{e^{s_{p}}}{\sum_{j}^{C} e^{s_{j}}} \right )
\end{align*}

where $C$ denotes the set of classes, $s$ the vector of predictions, and $s_p$ the prediction for the target class. 

The architecture of the CNN-LSTM model is composed of six layers. The input layer is followed by two convolutional layers, two LSTM~\cite{lstm} layers, and the output layer.

For evaluation, we used three different cross-validation techniques, namely, 
\begin{itemize}
    \item k-fold cross-validation on time windows of length 600 with $k=5$;
    \item leave-recordings-out cross-validation: One recording corresponds to starting and ending a recording in one go. In our study, the recordings are between \~{}46s and \~{}\num{1249}s, and we used an 80:20 train-test ratio. One recording can contain only one specific activity or multiple activities performed multiple times. This validation technique reflects the performance of the model when used in the app;
    \item lastly, we evaluate leave-one-subject-out cross-validation.
\end{itemize}

\subsubsection*{Optimal Sensor Placement}
\label{sub_sub_sec:osp}

Using the CNN-LSTM model, we trained 31 models for all sensor combinations and ranked them according to the $F1$ score. In addition, the results from the cross-validated feature metric $D_k$ were calculated from only 3000 data rows (300 s with 10 Hz) per activity and sensor. For the purpose of comparison of the trained models, we only included the results of $D_k$ for the same body locations using five sensors. The results for both approaches are shown in \autoref{tab:comparison_sensor_placement}. As can be seen, three of the four comparisons per number of sensors match the best-ranked sensor placement. There is only a slight difference in placing two sensors. Kendall's Tau coefficient $\tau$, a measure of the rank correlation between two variables, was calculated to evaluate the similarity between the rankings obtained from the CNN-LSTM model and $D_k$. A value of 1 indicates perfect agreement, while a value of -1 indicates perfect disagreement. The formula for Kendall's Tau coefficient is:

\begin{equation*}
\tau = \frac{2}{n(n-1)}\sum_{i<j}\mathrm{sgn}(x_i - x_j)\mathrm{sgn}(y_i - y_j)
\end{equation*}

where $n$ is the number of paired observations, $x$ and $y$ are the rankings of the two variables being compared, and $\operatorname{sgn}$ is the sign function.

\begin{table}[h!]
\footnotesize
  \begin{center}
    \caption{Comparison of Optimal Sensor Placement: Ground Truth vs. Predicted Placement. The table shows the sensor placement rankings based on the highest F1-score (Ground Truth) and the predicted placement rankings (Predicted Placement) for each number of IMUs (\#IMU). Kendall's tau coefficient is used to measure the correlation between the two rankings, where a coefficient of 1 indicates a perfect match. For \#IMU=1 and \#IMU=4, the predicted placement rankings are identical to the ground truth rankings, resulting in a Kendall's tau coefficient of 1. For \#IMU=2 and \#IMU=3, the predicted placement rankings show moderate correlation with the ground truth rankings, resulting in a Kendall's tau coefficient of 0.6 for both cases.}
    \label{tab:comparison_sensor_placement}
    \begin{tabularx}{\linewidth}
    {ccYYc}
      \toprule
      \textbf{\# IMU} & \textbf{Rank} &
      \textbf{Sensor Placement according to F1} & \textbf{$\boldsymbol{D_k}$} & \textbf{$\boldsymbol{\tau}$}\\
      \midrule
      \multirow{3}{*}{{\huge 1}} & 1 & \textcolor{teal}{LW}\tikzmark{11} & \tikzmark{12}\textcolor{teal}{LW} & \multirow{3}{*}{{\Large 1.0}}\\
      & 2 & \textcolor{purple}{RW}\tikzmark{21} & \tikzmark{22}\textcolor{purple}{RW} &\\
      & 3 & \textcolor{violet}{LF}\tikzmark{31} & \tikzmark{32}\textcolor{violet}{LF} &\\
      \midrule
      \multirow{3}{*}{{\huge 2}} & 1 & \textcolor{teal}{\{LW, PE\}}\tikzmark{41} & \tikzmark{42}\textcolor{violet}{\{LW, RW\}} & \multirow{3}{*}{{\Large 0.6}}\\
      & 2 & \textcolor{purple}{\{RW, PE\}}\tikzmark{51} & \tikzmark{52}\textcolor{teal}{\{LW, PE\}} &\\
      & 3 & \textcolor{violet}{\{LW, RW\}}\tikzmark{61} & \tikzmark{62}\textcolor{purple}{\{RW, PE\}} &\\
      \midrule
      \multirow{3}{*}{{\huge 3}} & 1 & \textcolor{teal}{\{LW, RW, PE\}}\tikzmark{71} & \tikzmark{72}\textcolor{teal}{\{LW, RW, PE\}} & \multirow{3}{*}{{\Large 0.6}}\\
      & 2 & \textcolor{purple}{\{LW, RW, LF\}}\tikzmark{81} & \tikzmark{82}\textcolor{violet}{\{LW, RW, RF\}} &\\
      & 3 & \textcolor{violet}{\{LW, RW, RF\}}\tikzmark{91} & \tikzmark{92}\textcolor{purple}{\{LW, RW, LF\}} &\\
      \midrule
      \multirow{3}{*}{{\huge 4}} & 1 & \textcolor{teal}{\{LW, RW, PE, RF\}}\tikzmark{101} & \tikzmark{102}\textcolor{teal}{\{LW, RW, PE, RF\}} & \multirow{3}{*}{{\Large 1.0}}\\
      & 2 & \textcolor{purple}{\{LW, RW, PE, LF\}}\tikzmark{111} & \tikzmark{112}\textcolor{purple}{\{LW, RW, PE, LF\}} &\\
      & 3 & \textcolor{violet}{\{LW, PE, LF, RF\}}\tikzmark{121} & \tikzmark{122}\textcolor{violet}{\{LW, PE, LF, RF\}} &\\
      \bottomrule
    \end{tabularx}
  \end{center}
\end{table}

\subsubsection*{Multimodal Activity Recognition}
\label{sub_sub_sec:mar}

\autoref{tab:multimodal_results} displays the results for each modality combination in the nursing dataset. Combining IMU and pose estimation data always performs best, whereas pose estimation data only always performs the worst.

\begin{table}[h!]
\footnotesize
  \begin{center}
    \caption{Classification results obtained for three different input data modalities: IMU, Pose Estimation, and the combination of IMU and Pose Estimation (IMU+Pose Estimation), using three evaluation methods: k-fold, leave-recordings-out, and leave-one-subject-out. The experiments were conducted to classify high-level activities constituting complex activities from nursing.
    } 
    \label{tab:multimodal_results}
    \begin{tabularx}{\linewidth}{cY>{\color{darkgray!60}}YY>{\color{darkgray!60}}YY>{\color{darkgray!60}}Y}
      \toprule
      \multirow{2}{*}{\textbf{Input Data}} & \multicolumn{2}{>{\hsize=\dimexpr2\hsize+2\tabcolsep+\arrayrulewidth\relax}Y}{\textbf{k-fold}} & \multicolumn{2}{>{\hsize=\dimexpr2\hsize+2\tabcolsep+\arrayrulewidth\relax}Y}{\textbf{leave-recordings-out}} & \multicolumn{2}{>{\hsize=\dimexpr2\hsize+2\tabcolsep+\arrayrulewidth\relax}Y}{\textbf{leave-one-subject-out}}\\
      & \textbf{Accuracy} & \textbf{F1} & \textbf{Accuracy} & \textbf{F1} & \textbf{Accuracy} & \textbf{F1}\\
      \midrule
      IMU & 0.823 $\pm 0.03$ & 0.822 $\pm 0.04$& 0.615 $\pm 0.04$ & 0.611 $\pm 0.06$ & 0.489 $\pm 0.05$ & 0.474 $\pm 0.06$\\
      Pose Estimation & 0.421 $\pm 0.07$ & 0.361 $\pm 0.03$ & 0.424 $\pm 0.06$ & 0.410 $\pm 0.07$ & 0.359 $\pm 0.05$ & 0.281 $\pm 0.08$\\
      IMU + Pose Estimation & \textbf{0.838 $\pm 0.01$} & \textbf{0.836 $\pm 0.02$} & \textbf{0.641 $\pm 0.01$} & \textbf{0.638 $\pm 0.03$}& \textbf{0.501 $\pm 0.02$} & \textbf{0.478 $\pm 0.03$}\\
      \bottomrule
    \end{tabularx}
  \end{center}
\end{table}

\section{Discussion}
\label{sec:discussion}

This paper addresses the development of a method for optimal sensor placement and a multimodal classification approach.

The cross-validated feature metric $D_k$ represents a suitable approach for the optimal determination of sensor localization. The approach seems to have recognized the importance of hand movements well. Similarly, multiple sensor combinations work correctly. Notably, this is the case even when the additional sensor detects relatively little motion, as for the pelvis. These results are in agreement with those obtained by the trained model. This could be explained by the fact that the sensors act as counterparts, one constituting a root point or reference point. Since different ML models can lead to different results, it is also difficult to conclude whether the minimal difference in the two sensors might be related to the used model.

The multimodal approach from IMUs and pose estimation data leads to increased classification accuracy overall. Nonetheless, the performance boost is not significant. There are two likely causes for this. (1) The results could be attributed to the different camera angles in data acquisition. A recording taken from the side lets the keypoints move closer together in 2D, which makes classification harder. The viewing angle, thus, plays an important role. (2) The lack of pose estimations under certain conditions. Pose estimations are not feasible when the camera does not capture the entire body or large portions. Out of the 51 recordings, pose estimation data is missing for ten.

\subsection{Limitations}
\label{sub_sec:limitations}

Our approach comes with some limitations. When forming pose estimations, distortions in the image can occur quickly if there are objects in front of the person or if the focus is shifted. This leads to low confidence values and, thus, gaps in data collection. Consequently, this would corrupt both a multimodal approach and the optimal determination of sensor positions. Furthermore, we use a 2D pose estimation approach that does not map depths. The missing dimension leads to an inaccurate distance representation of the observed person when the person turns or the recording angle changes. Our sensor placement optimization method is effective, straightforward to implement, and quick in execution, making it a practical choice for many applications. However, it's important to note that our study did not include a comparison with other sensor placement optimization methods. This was due to the lack of readily available implementations of alternative methods.

\section{Conclusion and Outlook}
\label{sec:conclusion}

The aim of the present research was to design a novel, lightweight optimal sensor placement approach. We make several contributions with our approach. First, the pose estimation technique is able to effectively anonymize test subjects. 
Second, we demonstrate the possibility of determining the optimal sensor placement without the necessity of actual IMUs. Videos from targeting activities are sufficient to determine the optimal placement. Lastly, the possibility to infer a multimodal classification approach.

Further improvement could be achieved by integrating a 3D pose estimation model, video recordings, and diverse sensor types. Future work will also aim to address implementing and comparing other sensor optimization methods.

\section*{Code \& Data Availability}

The study was conducted under subject's consent and ethical approval from the University of Potsdam, reference number 51/2021. The data from the feasibility study is accessible via Nextcloud~\cite{lab_data}. The code for the application including all used models is shared on GitHub~\cite{sonar}.

\bibliographystyle{ACM-Reference-Format}
\bibliography{sample-base}

\end{document}